\pdfoutput=1

\documentclass[11pt]{article}

\usepackage[final]{acl}

\usepackage{times}
\usepackage{latexsym}

\usepackage[T1]{fontenc}

\usepackage[utf8]{inputenc}

\usepackage{microtype}

\usepackage{inconsolata}

\usepackage{graphicx}

\usepackage{dirtytalk}
\usepackage[shortlabels,inline]{enumitem}
\usepackage{tikz,lipsum}
\usepackage[most]{tcolorbox}
\usepackage{ragged2e}
\usepackage{amsmath}
\usepackage{booktabs}
\usepackage{tabularray}
\usepackage{arydshln}
\usepackage{stmaryrd}
\usepackage{marvosym}
\usepackage{makecell}
\usepackage{multirow}
\usepackage{longtable}
\usepackage{tikz}
\usepackage{amssymb}
\usepackage{subcaption}
\usepackage{graphicx}
\usepackage{caption}

\usepackage{cleveref}
\crefname{section}{\S}{\S}
\crefname{table}{Table}{Tables}
\crefname{figure}{Fig.}{Figs.}
\crefname{algorithm}{Alg.}{}
\crefname{ALC@unique}{Line}{Lines}
\crefname{equation}{Eq.}{Eqs.}
\crefname{appendix}{App.}{Apps.}
\crefformat{section}{\S#2#1#3} 
\usepackage{soul}


\usepackage{colortbl}

\definecolor{tablegray}{RGB}{223, 242, 252}

\usepackage{todonotes}

\usepackage{xspace}
\newcommand{\DataReleaseURL}[0]{\url{https://github.com/dipta007/De-Presuppose}}

\title{ %
If We May De-Presuppose: Robustly Verifying Claims through Presupposition-Free Question Decomposition
}

\author{
\textbf{Shubhashis Roy Dipta, Francis Ferraro} \\
  Department of Computer Science and Electrical Engineering\\
  University of Maryland Baltimore County\\
  Baltimore, MD 21250 USA \\
  \texttt{\{sroydip1,ferraro\}@umbc.edu} \\
}

\begin{document}
\maketitle

\begin{abstract}


Prior work has shown that presupposition in generated questions can introduce unverified assumptions, leading to inconsistencies in claim verification. Additionally, prompt sensitivity remains a significant challenge for large language models (LLMs), resulting in performance variance as high as \textbf{3–6\%}. While recent advancements have reduced this gap, our study demonstrates that prompt sensitivity remains a persistent issue. To address this, we propose a structured and robust claim verification framework that reasons through presupposition-free, decomposed questions. Extensive experiments across multiple prompts, datasets, and LLMs reveal that even state-of-the-art models remain susceptible to prompt variance and presupposition. Our method consistently mitigates these issues, achieving up to a \textbf{2–5\%} improvement.\footnote{\DataReleaseURL}
\end{abstract}
\section{Introduction}
\begin{figure}
    \centering
    \includegraphics[width=1\linewidth]{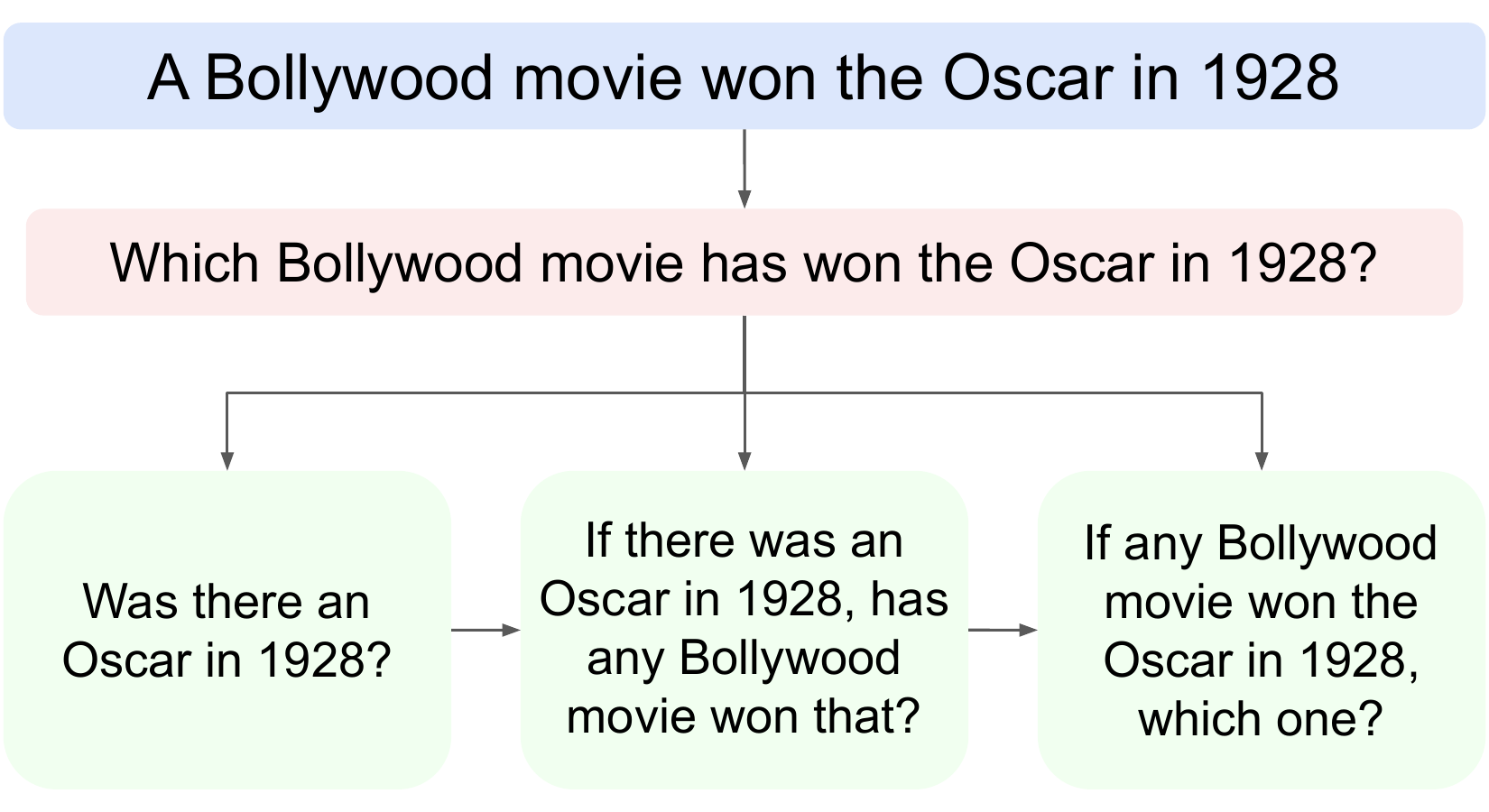}
    \caption{In contrast to simple claim decomposition, which can presuppose some parts of the claim as true (i.e. there were Oscars in 1928), our method automatically generates a collection of presupposition-free questions to verify each part of the claims. Here, it first asks whether there were Oscars in 1928, conditioning any subsequent questions on that answer.
    }
    \label{fig:example}
\end{figure}

While current large language models (LLMs) \citep{dubey_llama_2024, touvron_llama_2023, deepseek-ai_deepseek-r1_2025, qwen_qwen25_2025} demonstrate strong performance in claim verification \citep{tang_minicheck_2024, kamoi_wice_2023} when provided with ground truth evidence, they can improperly presuppose parts of the claim, which can then lead to incorrect conclusions or explanations. 
Consider, for example, the claim illustrated in \cref{fig:example}: ``\textit{A Bollywood movie won the Oscar in 1928}'': an LLM might simply verify whether ``\textit{any Bollywood movie won the 1928 Oscar},'' thereby accepting the existence of the 1928 Oscar without verifying it first.\footnote{The Oscars were first awarded in 1929.} In contrast, a more skeptical verifier would first ask, ``\textit{Was there an Oscar in 1928?}'' -- and only if that is true, proceed to verify the rest of the claim. This multi-layered reasoning highlights the need for decomposition-based verification that explicitly questions and validates every presupposition before drawing a conclusion.

We propose such a method of question-based decomposition and verification.
Specifically, \textbf{we demonstrate how presupposition-free questions can reduce prompt sensitivity, improve verification performance, and constrain overthinking}. This simple yet effective method creates structured reasoning paths that reduces prompt sensitivity by \textbf{2–5\%}, offering a significant robustness gain.
Our approach improves performance across both general and domain-specific scientific claim verification while reducing the need for labor-intensive prompt-tuning across datasets.

To summarize, our contributions are as follows:
\begin{enumerate}
\setlength{\itemsep}{0pt}
    \item We propose a decomposition and de-presupposition based question generation method to produce fine-grained questions.
    \item We show that while LLMs are highly sensitive to prompt variations, our structured reasoning approach significantly reduces this sensitivity, improving robustness across prompts.
    \item We demonstrate that our automated question generation achieves approximately 89\% coverage of atomic subclaims in a zero-shot setting.
    \item We show that our method outperforms strong baselines by 2–5\% across two datasets, three prompt variants, and three reasoning models.
\end{enumerate}

Our code is available at \DataReleaseURL.

\section{Method}

Given a claim (``\textit{Bollywood movie has won the Oscar in 1928}''), our approach proceeds in three steps: First, we automatically \textbf{decompose the claim into simpler questions} (\cref{sec:ques_decompostion}) -- ``\textit{Which Bollywood movie has won the Oscar in 1928?}''. We recognize that the generated questions may presuppose certain facts, such as ``\textit{There was an Oscar award in 1928}.'' To account for this, we next \textbf{reformulate and expand these questions to remove presuppositions}, resulting in presupposition-free questions (\cref{sec:ques_depresupposition}) -- ``(a) \textit{Was there an Oscar in 1928?}, (b) \textit{If there was, has any Bollywood movie won that?} (c) \textit{If yes, which one was that?}'' %
Finally, we use a reasoner to \textbf{verify the claim (\cref{sec:reason_through_ques}) with the help of these questions}. Our results show that this structuring improves claim verification performance.

\subsection{Question Generation} \label{sec:ques_decompostion}
\citet{chen-etal-2022-generating} demonstrated that using evidence during question generation yields significantly better results.
Motivated by this, we incorporate both claim and evidence in our question generation process. Inductively, we prompt an LLM to decompose a claim into separate, independent questions\footnote{Prompt is provided on \cref{prompt:que_decomposition}}. We adopted the prompt from \citet{kamoi_wice_2023} to better align with our question decomposition module instead of their subclaim decomposition. 
Specifically, their few-shot prompt includes examples of claim-to-subclaim decomposition, which we adapted by converting the assertive statements into questions.

\subsection{Question De-Presupposition} \label{sec:ques_depresupposition}
Through manual review of the generated questions, we identified that some of them presuppose information that may or may not be true, i.e., ``\textit{Which Bollywood movie has won the 1928 Oscar?}''.
While a model could, in theory, decompose the claim into presupposition-free questions, in practice we noticed this did not consistently happen. %
Rather, we noticed that models' subsequent reasoning over those questions would go a different way than intended. %
To address this, we employ LLM-based prompting to process the questions and decompose any presuppositions into multiple sub-questions, as illustrated in \cref{fig:example}. Specifically, for each question, we prompt the LLM to decompose it into presupposition-free atomic question\footnote{Prompt is provided on \cref{prompt:question_depresupposition}}.

\subsection{Reasoning through Questions} \label{sec:reason_through_ques}
Next, we employ a reasoning model that utilizes both the questions and the evidence to verify the claim. The key in our approach is that the presupposition-free questions from the previous step can guide the LLM to reason effectively, instead of directly verifying the claim based solely on the evidence. 
We do this reasoning by prompting an LLM; however, \citet{sclar_quantifying_2024} found that even large models are susceptible to prompt variance. Similarly, we found in our experiments that there was significant variation in performance based on which exact prompt we used: one was adopted from \citet{tang_minicheck_2024}'s verification prompt (``MiniCheck''), and two others were written by this paper's authors in an attempt to provide more \textbf{S}tructured \textbf{G}uidance (we call them ``SG1'' and ``SG2''). We provide full details and results in appendix (\cref{app:effect_of_prompt}).
Overall, we found that just by using these prompts alone performance varied by up to 6\% accuracy. This shows the need to mitigate prompt sensitivity; as our results will show, our method provides this ability.

\section{Experiments}


\paragraph{Evaluation Metric:}
We used balanced accuracy (BAcc) as the evaluation metric~\cite{tang_minicheck_2024, kamoi_wice_2023}, due to the imbalance between all the labels. To evaluate our question module, we have proposed a question coverage metric to assess the accuracy of generated questions (\cref{sec:rq4}).

\paragraph{Datasets:}
We used the BioNLI \cite{bastan_bionli_2022} and WiCE \cite{kamoi_wice_2023} dataset. While BioNLI features highly complex and domain specific scientific claims, WiCE has the real-life claims from Wikipedia. We have also curated a random balanced subset (300 claim-evidence pairs) of the BioNLI dataset to experiment with much costlier models, i.e., \texttt{o4-mini}. We call it BioNLI-300. Following previous works \cite{tang_minicheck_2024, kamoi_wice_2023}, we converted the WiCE dataset from a three-class problem to binary classification by considering both `Refuted' and `Partially Supported' as the `Refuted.' Detailed dataset statistics are in the appendix (\cref{tab:dataset_stat}). 

\paragraph{Experimental Setup:}
We have used \texttt{Qwen/QwQ-32B} and \texttt{o4-mini} as the reasoner models. Both models were run with the default temperature settings, while for \texttt{o4-mini}, we utilized the ``high'' reasoning setup. More details on the implementation is provided on \cref{app:exp_setup}.

\subsection{RQ1: Does de-presupposition help?} \label{sec:rq3}
\begin{table}[!t]
\resizebox{\columnwidth}{!}{%
    \begin{tabular}{@{}lccc@{}}
    Prompt              & \thead{Only-Reasoner}     & \thead{Reasoner + Question \\Decomposition} & \thead{Our Method}        \\ 
    
    \midrule
    \multicolumn{4}{@{}l}{\textbf{\textit{BioNLI-FULL}}} \\
    SG2 & 73.74 ± 0.10 & \textbf{76.57 ± 0.08} & \textbf{76.57 ± 0.19} \\
    SG1 & 72.34 ± 0.26 & 76.72 ± 0.14 & \textbf{77.73 ± 0.06} \\
    MiniCheck           & 77.58 ± 0.14 & 78.04 ± 0.14 & \textbf{78.32 ± 0.15} \\

    \midrule
    \multicolumn{4}{@{}l}{\textbf{\textit{BioNLI-300}}} \\
    SG2 & 69.11 ± 0.42 & 72.56 ± 1.50 & \textbf{73.44 ± 1.40} \\
    SG1 & 68.44 ± 0.68 & 74.11 ± 1.29 & \textbf{75.00 ± 2.18} \\
    MiniCheck           & 73.33 ± 0.98 & 74.33 ± 1.25 & \textbf{75.11 ± 1.10} \\
    
    \midrule
    \multicolumn{4}{@{}l}{\textbf{\textit{WiCE}}} \\
    SG2 & 76.36 ± 0.27 & \textbf{79.32 ± 0.49} & 79.03 ± 0.00 \\
    SG1 & 73.41 ± 0.56 & \textbf{78.23 ± 0.64} & 76.42 ± 0.92 \\
    MiniCheck           & 80.70 ± 0.30 & 81.72 ± 0.50 & \textbf{82.25 ± 0.62} \\

    \bottomrule
    \end{tabular}
}
\caption{Results (with standard deviation) on the BioNLI and WiCE dataset. Full results are reported in the appendix (\cref{tab:our_method_results}).
}
\label{tab:does_dep_help}
\end{table}

We consider three setups: (1) Only Reasoner, (2) Reasoner + Question Decomposition and (3) Our method with de-presupposition. The results of all the settings are reported in \cref{tab:does_dep_help}. %
The findings indicate that de-presupposition consistently improves performance across all settings. The full results in \cref{tab:our_method_results} also show that most notable gains are observed when paired with a more capable reasoning model, such as \texttt{o4-mini}. We hypothesize that less capable models may not fully leverage the benefits of presupposition-free questions. The impact of de-presupposition is more pronounced on complex, multi-hop datasets like BioNLI, while the improvements are less substantial on simpler datasets such as WiCE.

We also experimented with adding an explicit answer module to further ease the burden on the verifier. However, as shown in \cref{tab:explicit_answer_module}, this addition often degrades performance. We hypothesize that generating intermediate answers may introduce errors that propagate to the verifier. While a similar risk exists for question decomposition and de-presupposition, breaking a claim into sub-questions—and refining them into presupposition-free forms—is comparatively easier and less error-prone than answering questions directly from long, unstructured documents.

\subsection{RQ2: Do we cover sub-claims?} \label{sec:rq4}
To analyze how well our generated questions cover the critical parts of the claim, we use the decomposed subclaims from the WiCE dataset. We employed \texttt{Qwen/Qwen3-32B} to evaluate coverage. During evaluation, for each subclaim, we asked the model whether it was addressed by at least one question or a combination of multiple questions. This approach accounts for the fact that a single subclaim may be implicitly addressed by multiple questions, aligning with our goal of generating as many skeptical and granular questions as possible. For example, the claim ``San Jose is the biggest city in Texas'' can be decomposed into two questions: (1) \textit{Is San Jose a city in Texas?} and (2) \textit{If yes, is it the biggest?}
To avoid occasional contextualization issues when aligning questions with subclaims, %
we provided the ground truth evidence as additional context, allowing the LLM to properly de-contextualize the questions.\footnote{Prompt is provided in \cref{prompt:question_coverage}}

\begin{table}[!t]
    \centering
        \begin{tabular}{@{}lr@{}}
            \toprule
            Question Decomposer & Question Coverage \\
            \midrule
            o4-mini             & 89.16 ± 0.20      \\
            Qwen/QwQ-32B        & 87.41 ± 0.00      \\
            \bottomrule
        \end{tabular}
    \caption{Coverage of the sub-claims from the WiCE dataset. 
    Standard deviation is across 3 runs for question decomposing.}
    \label{tab:ques_coverage_wice}
\end{table}

\cref{tab:ques_coverage_wice} shows we achieve $\sim$89\% coverage of subclaims, regardless of the underlying LLM. These results underscore that question decomposition is highly effective in zero-shot settings, 
while the low standard deviation across runs highlights the robustness of the question generation.
\subsection{RQ3: Can we mitigate prompt variance?} \label{sec:rq2}
In the next RQ, we explore if our proposed method can mitigate the prompt sensitivity. To analyze this question, 
we used our method 
across two datasets -- WiCE \cite{kamoi_wice_2023} and BioNLI \cite{bastan_bionli_2022}, using three different prompts\footnote{Prompts are shown in \cref{prompt:question_generation_verbose_0}, \cref{prompt:question_generation_verbose_1}, \cref{prompt:modified_mini_check}.} and three different models (\texttt{o4-mini}, \textit{Qwen/QwQ-32B}, \texttt{Qwen/QwQ-32B}). Due to the high API cost of \texttt{o4-mini} and the large number of samples in the BioNLI dataset (\cref{tab:dataset_stat}), we used only the \texttt{Qwen} models\footnote{Due to space, \texttt{o4-mini} results on the BioNLI-300 dataset are provided in \cref{tab:our_method_results} in the appendix.} for evaluating the BioNLI full dataset.


\begin{figure}[!t]
    \centering
    \includegraphics[width=1\linewidth]{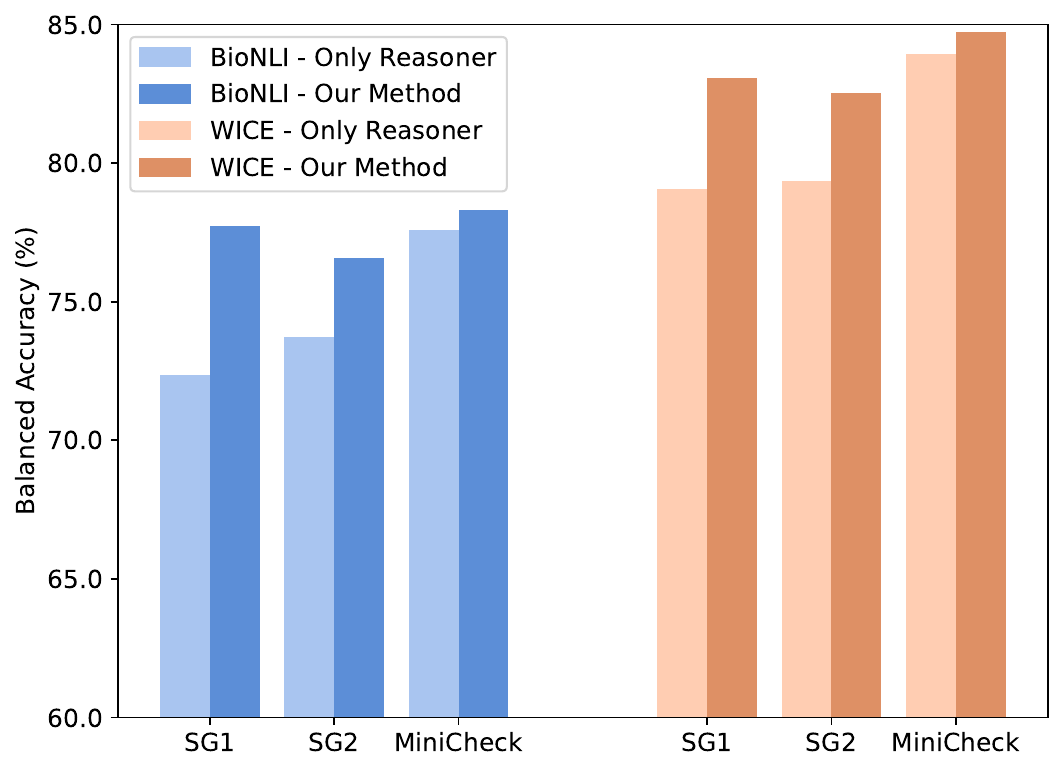}
    \caption{Balanced accuracy on the BioNLI (blue bar) and WiCE (orange bar) benchmarks under three prompt variants. Full results are reported in appendix (\cref{fig:our_method_plot_full}).}
    \label{fig:our_method_plot}
\end{figure}

The results presented in \cref{fig:our_method_plot} offer several key insights: (1) Our method outperforms the baseline across all datasets, prompts, and models, highlighting its effectiveness, and (2) It mitigates the performance degradation caused by prompt variance, demonstrating the robustness of the method.
\subsection{Error Analysis}
\begin{figure}[t]
    \centering
    \includegraphics[width=1\linewidth]{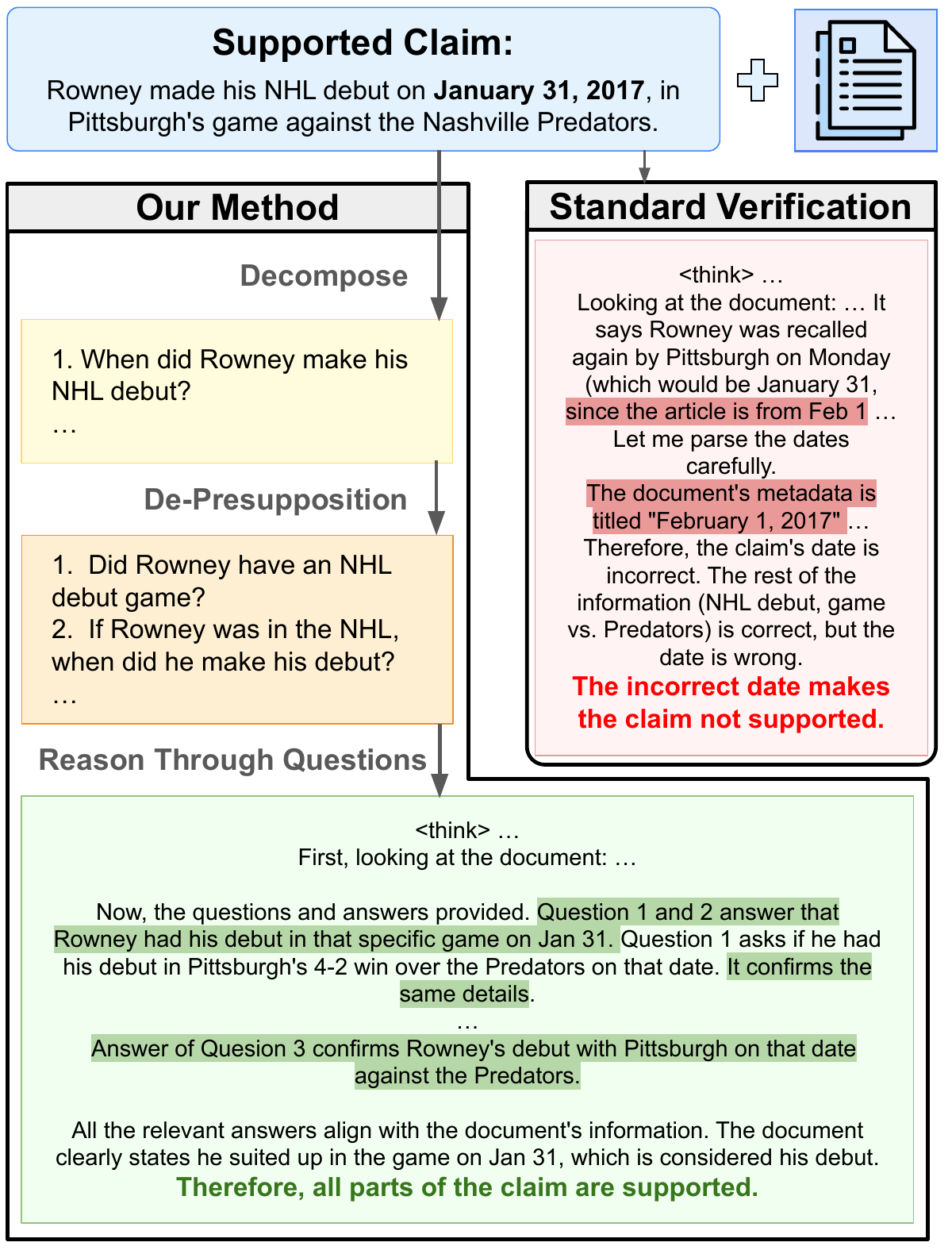}
    \caption{Question-answer-based reasoning (left) vs. typical verification by a reasoner model (right) on a \textsc{Supported} claim from WiCE \citep{kamoi_wice_2023}. 
    }
    \label{fig:intro}
\end{figure}

In \cref{fig:intro} we present an excerpted annotated example from WiCE. On the left we show our approach, and on the right we show an LLM reasoner without our approach. Note that our structured reasoning can mitigate overthinking. In the red box, \texttt{Qwen3-32B} over-analyzed the evidence, placing unnecessary emphasis on the publication date (Feb 1) instead of the key event date (Jan 31), leading to an incorrect conclusion. In contrast, our method first decomposes the claim into high-level questions and then further into subatomic questions through de-presupposition, filtering out irrelevant information. We show additional full examples in the \cref{app:qual_analysis}. 

\section{Related Works}


Many works have explored claim verification \cite{kamoi_wice_2023, tang_minicheck_2024, zha_alignscore_2023, min_factscore_2023, song_veriscore_2024, wang_openfactcheck_2024}, but the decomposition of claims into questions is a relatively newer field. \citet{chen-etal-2022-generating} employed a trained question decomposer model to break down claims into multiple questions. In contrast, our method focuses on zero-shot question decomposition using the inherent knowledge embedded in LLMs, allowing it to be applied across different domains of claim verification, such as scientific, real-life, and political claims.

\citet{hu-etal-2025-decomposition} demonstrated that direct claim decomposition yields mixed results depending on the strength of the verifier. Our findings are similar, with performance varying based on the type of prompts used. However, our work focuses on reducing this issue to a greater extent. Similarly, \citet{fleischer_square_2025} employed a question-based decomposition to answer the main question.
\citet{xue_decompose_nodate} applied a similar question-answer graph with a voting mechanism to solve mathematical problems. However, claim verification requires more extensive reasoning because its questions are less deterministic—unlike mathematical questions, claims do not follow a logical rationale, and thus the decomposition requires more filtering and reasoning.

In the multimodal domain, \citet{cho2023davidsonian} introduced a method of decomposing text into a graph of questions to evaluate text-to-image models. Similarly, \citet{jiang_diem_2024} decomposed both questions and images to answer the main question.

Recently, \citet{11012707} has proposed a pipeline through question-answering to detect if a text has presupposition or not. Similarly, \citet{kim2021linguist} has shown that using presupposition-based decomposition improve the retrieval and hence improve the end-QA performance.

In contrast, our method focuses on a less explored aspect of decomposition, presupposition, and its impact on prompt variance and overthinking in reasoning models using dynamic question generation.

\section{Conclusion}
We introduce a novel question decomposition approach for generating presupposition-free, atomic questions that systematically interrogate each part of a claim. This study is inspired by how humans think by decomposing and verifying each part of a complex claim. Using our approach, we have shown that we can mitigate the prompt sensitivity by constraining it to thinking structurally than free form. We hope this line of research encourages further exploration into decomposition-based reasoning as a foundation for building more trustworthy and transparent claim verification systems.

\section*{Limitations}
While the current LLMs are very effective as LLM-as-a-judge, there is always an inherent nature of randomness which can affect the question coverage metric. Also, we have not tested different other prompting methods due to the size of the dataset, computation constraints and high API cost. Furthermore, while we have shown empirically that our method improves the end-performance on claim verification task, we have not manually verified the outputs (if they are presupposition free) due to the size of the dataset.

\section*{Acknowledgments}
We thank the reviewers for their detailed comments and suggestions. %
Some experiments were conducted on the UMBC HPCF, supported by the National Science Foundation under Grant No. CNS-1920079. %
This material is also based on research that is in part supported by the Army Research Laboratory, Grant No. W911NF2120076, and by DARPA for the SciFy program under agreement number HR00112520301. %
The U.S. Government is authorized to reproduce and distribute reprints for Governmental purposes notwithstanding any copyright notation thereon. The views and conclusions contained herein are those of the authors and should not be interpreted as necessarily representing the official policies or endorsements, either express or implied, of DARPA or the U.S. Government.


\bibliography{custom}

\appendix
\section{Appendix}
\label{sec:appendix}

\subsection{Prompts} \label{app:prompts}
\subsubsection{Question Decomposition} \label{prompt:que_decomposition}
\prompt[
{}
]{
You are given a claim, your task is to decompose it into multiple independent and individual questions. DON'T generate any other text than the questions. You are given some examples below and the input claim at the end. \\

Claim: Other title changes included Lord Steven Regal and The Nasty Boys winning the World Television Championship and the World Tag Team Championship respectively. \\
Questions: \\
- Did Lord Steven Regal win the World Television Championship? \\
- Did The Nasty Boys win the World Tag Team Championship? \\

Claim: The parkway was opened in 2001 after just under a year of construction and almost two decades of community requests. \\
Questions: \\
- When was the parkway opened? \\
- How long was the construction period for the parkway? \\
- How many years of community requests preceded the opening of the parkway? \\

Claim: Touring began in Europe in April–June with guitarist Paul Gilbert as the opening act, followed by Australia and New Zealand in July, Mexico and South America in late July–August, and concluding in North America in October–November. \\
Questions: \\
- When did touring begin in Europe? \\
- Who was the opening act during the touring in Europe? \\
- Which months covered the Australia tour? \\
- Which months covered the New Zealand tour? \\
- Which months covered the Mexico tour? \\
- Which months covered the South America tour? \\
- Which months covered the North America tour? \\
- Where did the touring conclude? \\

Claim: In March 2018, the company partnered With Amazon Web Services (AWS) to offer Al-enabled conversational solutions to customers in India. \\
Questions: \\
- When did the company partner with AWS? \\
- What was the purpose of the partnership? \\

Claim: The most significant of these is in Germany, which now has a Yazidi community of more than 200,000 living primarily in Hannover, Bielefeld, Celle, Bremen, Bad Oeynhausen, Pforzheim and Oldenburg. \\
Questions: \\
- Which country hosts the largest Yazidi community? \\
- How large is the Yazidi community in Germany? \\
- In which cities are the Yazidi community in Germany primarily located? \\

Claim: A previous six-time winner of the Nations' Cup, Sebastian Vettel became Champion of Champions for the first time, defeating Tom Kristensen, who made the final for the fourth time, 2–0. \\
Questions: \\
- How many times had Sebastian Vettel won the Nations' Cup before? \\
- What title did Sebastian Vettel achieve for the first time? \\
- Whom did Sebastian Vettel defeat in the final? \\
- How many finals had Tom Kristensen reached? \\
- What was the final score between Sebastian Vettel and Tom Kristensen? \\

Claim: \promptvar{claim} \\
Questions: \\
}

\subsubsection{Question De-Presupposition} \label{prompt:question_depresupposition}
\prompt[
{}
]{
You are given a question that may contain presuppositions — assumptions that are implied but not necessarily true. Your task it to rewrite the question into one or more simpler de-contextualized questions that do not contain these presuppositions. DO NOT generate anything else other than the questions. You are also given some examples below and the input question at the end. \\

Question: \\
Which Bollywood movie has won the Oscar in 1928? \\

Rewritten questions: \\
- Was there an Oscar in 1928? \\
- If there was an Oscar in 1928, has any Bollywood movie won that? \\
- If any Bollywood movie won the Oscar in 1928, which one? \\

Question: \\
Which english movie was directed by Christopher Nolan? \\

Rewritten questions: \\
- Is Christopher Nolan a director? \\
- Has Christopher Nolan directed any english movie? \\
- If Christopher Nolan has directed any english movie, which one? \\

Question: \\
\promptvar{question} \\

Rewritten questions: \\
}

\subsubsection{Reasoner: SG1} \label{prompt:question_generation_verbose_0}
\prompt[
{}
]{
You are an AI model tasked with verifying claims using zero-shot learning. Your job is to analyze a given claim along with provided evidence (i.e. corpus articles) and decide whether the available evidence would likely support or not support the claim. You are also given some questions that can help you analyze the claim and evidence. \\

Claim to evaluate: \\
<claim> \\
\promptvar{CLAIM} \\
</claim> \\

Additional evidence provided: \\
<corpus\_text> \\
\promptvar{EVIDENCE} \\
</corpus\_text> \\

Questions to consider: \\
<questions> \\
\promptvar{QUESTIONS} \\
</questions> \\

Guidelines: \\
1. Evaluate the claim only based on the evidence provided. \\
2. It's possible that you are given multiple evidence articles. It is also possible that some of the evidence articles are not relevant to the claim. Use your best judgement to determine which evidence to use and which to ignore. \\
3. Consider answering the questions one by one, before making a final decision. \\
4. If relevant information is not present in the evidence, then it is possible that the claim is not supported by the evidence. Use your best judgement and previous knowledge to make a decision. \\

After your analysis, output exactly one JSON object with exactly two keys: "reasoning" and "decision". The value associated with "decision" must be exactly one word – either "SUPPORTED" or "NOT\_SUPPORTED" (uppercase, with no additional text). Do not add any markdown formatting, code fences, or additional text. " \\
"The output must start with an opening curly brace \{\{ and end with a closing curly brace \}\}. \\

Example output format: \\
\{\{"reasoning": "Your brief explanation here (one or two sentences).", "decision": "SUPPORTED or NOT\_SUPPORTED"\}\} \\

Now, please evaluate the above claim. \\
}

\subsubsection{Reasoner: SG2} \label{prompt:question_generation_verbose_1}
\prompt[{}]{
You are an AI model tasked with verifying claims using zero-shot learning. Your job is to analyze a given claim along with provided evidence and decide whether the available evidence would likely support or not support the claim. You are also given some questions that can help you analyze the claim and evidence. \\

\#\#\# Instructions: \\
1. Evaluate the claim only based on the evidence provided. \\
2. Consider answering the questions one by one, before making a final decision. \\
3. It is possible that some of the questions are not relevant to the claim. Use your best judgement to determine which questions to answer and which to ignore. \\
4. Finally, analyze the claim, questions and evidence together and determine the label that best describes the relationship between the claim and the evidence. \\
5. The meaning of different labels: \\
    - SUPPORTED: The claim is supported by the evidence. \\
    - NOT\_SUPPORTED: The claim is not supported by the evidence. \\

\#\#\# Output Format: \\
After your analysis, output exactly one JSON object with exactly two keys: "reasoning" and "decision". The value associated with "decision" must be exactly one word – either "SUPPORTED" or "NOT\_SUPPORTED" (uppercase, with no additional text). Do not add any markdown formatting, code fences, or additional text. " \\
"The output must start with an opening curly brace \{\{ and end with a closing curly brace \}\}. \\

Example output format: \\
\{\{"reasoning": "Your brief explanation here (one or two sentences).", "decision": "SUPPORTED or NOT\_SUPPORTED"\}\} \\

\#\#\# Input: \\
Evidence: \\
\promptvar{EVIDENCE} \\

Claim to evaluate: \\
\promptvar{CLAIM} \\

Questions to consider: \\
\promptvar{QUESTIONS} \\

\#\#\# Output: \\
}

\subsubsection{Reasoner: MiniCheck} \label{prompt:modified_mini_check}

\prompt[
{}
]{
Determine whether the provided claim is supported by the corresponding document. You are also given some decomposed questions derived from the claim. Reason through the questions to support your judgment. Support in this context implies that all information presented in the claim is substantiated by the document. If not, it should be considered not supported. Its possible that some of the questions are not relevant to the claim. Use your best judgement to determine which questions to consider and which to ignore. Fall back to the provided document when you are not sure about the question. \\

Document: \\
\promptvar{EVIDENCE} \\

Claim: \\
\promptvar{CLAIM} \\

Questions to consider: \\
\promptvar{QUESTIONS} \\

Please assess the claim's support with the document by responding with either "SUPPORTED" or "NOT\_SUPPORTED". Do not generate anything else other than the answer. \\

Answer: \\
}

\subsubsection{Question Coverage} \label{prompt:question_coverage}
To find out the coverage of the questions generated in the WiCE dataset, we have used the following prompt. The prompt has access to one subclaim and multiple questions and asks to find out if the subclaim is implicitly or explicitly covered by the questions. In the initial experiments, we have found out that given the evidence during the evaluation performs better due to the nominal references present in the subclaim or questions.
\prompt[
{}
]{
Given a claim, evidence, and a list of questions, analyze whether the questions collectively are sufficient to verify or refute the entire claim. \\

\#\#\# Instructions \\
- We are looking for coverage of the claim not completeness of the questions. So, if some questions are not relevant to the claim, that's fine. But if the relevant questions do not cover the whole claim, then the coverage is not good. \\
- The question does not need to ask the specific claim explicitly. If answering the question would verify the claim, then it covers the claim. \\
- It is possible that multiple questions together cover the claim. It is not necessary that the claim is covered by a single question. \\
- If a question and claim refer to similar, but non-identical concepts, use the provided evidence to determine whether the question and claim are referring to the same concept or not. For example, the claim may refer to "the machine learning technique," while the question may ask about "the supervised learning technique." Because the questions were generated based on the provided evidence, consider this evidence when determining your final answer. \\
- Begin by providing 1-2 sentences explaining your reasoning for the coverage of the claim. \\
- Afterward, output yes if the questions cover the claim completely, or no if they do not. \\
- Structure your final response into two sections: \\

        \quad- EXPLANATION: (your reasoning in 1-2 sentences) \\
        
        \quad- ANSWER: (Yes if the questions cover the claim completely, or No if they do not) \\

\#\#\# Evidence \\
\promptvar{evidence} \\

\#\#\# Claim \\
\promptvar{claim} \\

\#\#\# Questions \\
\promptvar{questions} \\
}







    




\subsection{Experimental Setup} \label{app:exp_setup}

We have used \texttt{Qwen/QwQ-32B} and \texttt{o4-mini} as the reasoner models. For the QwQ model, we have used VLLM \cite{kwon2023efficient} for inference (one H100) and the OpenAI API.
Both models were run with the default temperature settings, while for \texttt{o4-mini}, we utilized the ``high'' reasoning setup. We performed three runs to report the mean performance with standard deviation for the BioNLI-300 and WiCE datasets. We conducted two runs for the BioNLI full dataset and \texttt{o4-mini} results due to high computation time and API costs.
\subsection{Effect of Prompt Variance in Claim Verification} \label{app:effect_of_prompt}

\begin{table}[!t]
    \centering
    \resizebox{\columnwidth}{!}{
        \begin{tabular}{@{}llcc@{}}
        Prompt & BAcc         & Supported          & Refuted  \\
        \midrule
        

        \multicolumn{4}{@{}l}{\textbf{\textit{BioNLI-300}}} \\
        SG2  & 70.83 \textcolor{red}{(-3.50)} & 87.33 & 54.33 \\
        SG1  & 68.33 \textcolor{red}{(-6.00)} & 86.67 & 50.00 \\
        MiniCheck   & 74.33 & 74.00 & 74.67 \\
        \midrule
        \multicolumn{4}{@{}l}{\textbf{\textit{WiCE}}} \\
        SG2  & 79.35 \textcolor{red}{(-4.60)} & 91.89 & 66.80 \\
        SG1  & 79.08 \textcolor{red}{(-4.87)} & 94.59 & 63.56 \\
        MiniCheck   & 83.95 & 82.88 & 85.02 \\
        
        \bottomrule
        \end{tabular}
    }
    \caption{Comparison of different prompts -- SG1 -  \cref{prompt:question_generation_verbose_0}, SG2 -  \cref{prompt:question_generation_verbose_1}, MiniCheck - \cref{prompt:modified_mini_check}. \texttt{o4-mini} with high-reasoning setup was used as reasoner (Average score is reported over two runs)}
    \label{tab:prompt_compare}
\end{table}

\citet{sclar_quantifying_2024} found that even large models are susceptible to prompt variance, to confirm that on our task we have used 3 different prompts.

\paragraph{MiniCheck Prompt:} We have adopted the \citet{tang_minicheck_2024}'s verification prompt on this version. The exact prompt is reported on \cref{prompt:modified_mini_check}.

\paragraph{Structured Guidance:} To confirm the prompt variance, we have used 2 different structured guidance prompts with one minimal change -- reorganization of the sections. Both prompts are reported on \cref{prompt:question_generation_verbose_0} and \cref{prompt:question_generation_verbose_1}, both of those prompts were written by the authors and refined using the BioNLI validation set. We call them ``SG1'' and ``SG2'' respectively.

The results for different prompts are shown on the \cref{tab:prompt_compare}. The results show that the claim verification task, like many other tasks \citep{zhuo_prosa_2024}, is susceptible to the prompt (3-6\%). The difference between the SG1 and SG2 is more shocking, as those two prompts are basically the same, with some minor reorganization of sections. Our method can effectively mitigate this prompt sensitivity of a reasoner model.
\subsection{Additional Results}
The full result of the experiment described in \cref{sec:rq2} is shown on the \cref{tab:our_method_results}. In addition to the previous datasets, we have also provided results on the BioNLI-300 dataset to include the results for \texttt{o4-mini} as reasoner.

\begin{table*}[!ht]
\centering
\resizebox{0.9\textwidth}{!}{%
    \begin{tabular}{@{}lccc>{\columncolor{tablegray}}ccc@{}}
    Prompt              & Reasoner     & Question Decomposer & \thead{De-\\Presupposition} & BAcc         & Supported    & Refuted    \\

    \midrule
    \multicolumn{4}{@{}l}{\textbf{\textit{BioNLI-FULL}}} & \cellcolor{tablegray}\\ 
    SG2 & Qwen3-32B & Qwen3-32B & $\checkmark$ & 76.79 ± 0.16 & 90.37 ± 0.72 & 63.21 ± 0.39 \\
    SG2 & Qwen3-32B & Qwen3-32B & $\times$ & 75.92 ± 0.16 & 89.83 ± 0.09 & 62.00 ± 0.42 \\
    SG2 & Qwen3-32B &           -           & - & 74.10 ± 0.04 & 89.65 ± 0.27 & 58.54 ± 0.19 \\
    
    SG2 & QwQ-32B         & QwQ-32B         & $\checkmark$ & 76.57 ± 0.19 & 90.10 ± 0.27 & 63.04 ± 0.11 \\
    SG2 & QwQ-32B         & QwQ-32B         & $\times$ & 76.57 ± 0.08 & 89.92 ± 0.09 & 63.23 ± 0.08 \\
    SG2 & QwQ-32B         &             -         & - & 73.74 ± 0.10 & 91.49 ± 0.14 & 55.99 ± 0.06 \\
    
    SG1 & Qwen3-32B & Qwen3-32B & $\checkmark$ & 76.16 ± 0.01 & 87.49 ± 0.18 & 64.84 ± 0.15 \\
    SG1 & Qwen3-32B & Qwen3-32B & $\times$ & 75.38 ± 0.31 & 87.76 ± 0.72 & 63.00 ± 0.10 \\
    SG1 & Qwen3-32B &           -           & - & 71.49 ± 0.11 & 89.87 ± 0.41 & 53.10 ± 0.18 \\
    
    SG1 & QwQ-32B         & QwQ-32B         & $\checkmark$ & 77.73 ± 0.06 & 87.04 ± 0.00 & 68.41 ± 0.11 \\
    SG1 & QwQ-32B         & QwQ-32B         & $\times$ & 76.72 ± 0.14 & 87.98 ± 0.05 & 65.46 ± 0.32 \\
    SG1 & QwQ-32B         &           -           & - & 72.34 ± 0.26 & 93.92 ± 0.32 & 50.76 ± 0.20 \\
    
    MiniCheck  & Qwen3-32B & Qwen3-32B & $\checkmark$ & 77.60 ± 0.10 & 81.14 ± 0.14 & 74.05 ± 0.33 \\
    MiniCheck           & Qwen3-32B & Qwen3-32B & $\times$ & 76.34 ± 0.32 & 80.65 ± 0.54 & 72.03 ± 0.10 \\
    MiniCheck  & Qwen3-32B &            -          & - & 76.29 ± 0.28 & 80.11 ± 0.09 & 72.46 ± 0.48 \\
    
    MiniCheck  & QwQ-32B         & QwQ-32B         & $\checkmark$ & 78.32 ± 0.15 & 86.45 ± 0.14 & 70.18 ± 0.16 \\
    MiniCheck           & QwQ-32B         & QwQ-32B         & $\times$ & 78.04 ± 0.14 & 86.54 ± 0.32 & 69.54 ± 0.03 \\
    MiniCheck  & QwQ-32B         &            -          & - & 77.58 ± 0.14 & 84.92 ± 0.23 & 70.23 ± 0.06 \\

    \midrule
    \multicolumn{4}{@{}l}{\textbf{\textit{BioNLI-300}}} & \cellcolor{tablegray}\\ 
    SG2 & o4-mini      & o4-mini      & $\checkmark$ & 74.83 ± 0.50 & 85.00 ± 1.00 & 64.67 ± 0.00 \\
    SG2 & o4-mini      & o4-mini      & $\times$ & 71.33 ± 0.33 & 85.67 ± 0.33 & 57.00 ± 1.00 \\
    SG2 & o4-mini      & -            & - & 70.83 ± 0.17 & 87.33 ± 0.00 & 54.33 ± 0.33 \\
    
    SG2 & QwQ-32B & QwQ-32B & $\checkmark$ & 73.44 ± 1.40 & 85.56 ± 0.83 & 61.33 ± 1.96 \\
    SG2 & QwQ-32B & QwQ-32B & $\times$ & 72.56 ± 1.50 & 86.44 ± 1.13 & 58.67 ± 1.96 \\
    SG2 & QwQ-32B & -            & - & 69.11 ± 0.42 & 88.00 ± 0.54 & 50.22 ± 0.31 \\
    
    SG1 & o4-mini      & o4-mini      & $\checkmark$ & 73.00 ± 0.33 & 82.33 ± 1.00 & 63.67 ± 0.33 \\
    SG1 & o4-mini      & o4-mini      & $\times$ & 71.67 ± 0.00 & 83.33 ± 0.67 & 60.00 ± 0.67 \\
    SG1 & o4-mini      & -            & - & 68.33 ± 0.00 & 86.67 ± 0.00 & 50.00 ± 0.00 \\
    
    SG1 & QwQ-32B & QwQ-32B & $\checkmark$ & 75.00 ± 2.18 & 86.00 ± 0.54 & 64.00 ± 3.81 \\
    SG1 & QwQ-32B & QwQ-32B & $\times$ & 74.11 ± 1.29 & 86.44 ± 0.83 & 61.78 ± 1.75 \\
    SG1 & QwQ-32B & -            & - & 68.44 ± 0.68 & 92.44 ± 0.83 & 44.44 ± 0.63 \\
    
    MiniCheck           & o4-mini      & o4-mini      & $\checkmark$ & 74.67 ± 0.33 & 79.00 ± 1.00 & 70.33 ± 0.33 \\
    MiniCheck  & o4-mini      & o4-mini      & $\times$ & 71.83 ± 0.17 & 79.00 ± 0.33 & 64.67 ± 0.00 \\
    MiniCheck           & o4-mini      & -            & - & 74.33 ± 0.00 & 74.00 ± 0.00 & 74.67 ± 0.00 \\
    
    MiniCheck           & QwQ-32B & QwQ-32B & $\checkmark$ & 75.11 ± 1.10 & 82.44 ± 0.63 & 67.78 ± 1.57 \\
    MiniCheck  & QwQ-32B & QwQ-32B & $\times$ & 74.33 ± 1.25 & 83.33 ± 1.63 & 65.33 ± 0.94 \\
    MiniCheck           & QwQ-32B & -            & - & 73.33 ± 0.98 & 82.67 ± 1.09 & 64.00 ± 1.44 \\
    
    \midrule
    \multicolumn{4}{@{}l}{\textbf{\textit{WiCE}}} & \cellcolor{tablegray}\\ 
    SG2 & o4-mini      & o4-mini             & $\checkmark$ & 82.54 ± 1.46 & 90.99 ± 0.90 & 74.09 ± 2.02 \\
    SG2 & o4-mini      & o4-mini      & $\times$ & 83.32 ± 0.17 & 92.34 ± 1.35 & 74.29 ± 1.01 \\
    SG2 & o4-mini      & -                   & - & 79.35 ± 0.00 & 91.89 ± 0.00 & 66.80 ± 0.00 \\
    
    SG2 & QwQ-32B & QwQ-32B        & $\checkmark$ & 79.03 ± 0.00 & 85.59 ± 0.00 & 72.47 ± 0.00 \\
    SG2 & QwQ-32B & QwQ-32B & $\times$ & 79.32 ± 0.49 & 84.68 ± 0.74 & 73.95 ± 1.38 \\
    SG2 & QwQ-32B & -                   & - & 76.36 ± 0.27 & 86.19 ± 0.42 & 66.53 ± 0.19 \\
    
    SG1 & o4-mini      & o4-mini             & $\checkmark$ & 83.07 ± 0.53 & 91.44 ± 0.45 & 74.70 ± 0.61 \\
    SG1 & o4-mini      & o4-mini      & $\times$ & 83.52 ± 0.33 & 92.34 ± 0.45 & 74.70 ± 0.20 \\
    SG1 & o4-mini      & -                   & - & 79.08 ± 0.00 & 94.59 ± 0.00 & 63.56 ± 0.00 \\
    
    SG1 & QwQ-32B & QwQ-32B        & $\checkmark$ & 76.42 ± 0.92 & 83.48 ± 1.12 & 69.37 ± 1.99 \\
    SG1 & QwQ-32B & QwQ-32B & $\times$ & 78.23 ± 0.64 & 85.89 ± 0.42 & 70.58 ± 1.66 \\
    SG1 & QwQ-32B & -                   & - & 73.41 ± 0.56 & 91.89 ± 0.74 & 54.93 ± 0.69 \\
    
    MiniCheck  & o4-mini      & o4-mini             & $\checkmark$ & 84.74 ± 0.40 & 86.49 ± 0.00 & 83.00 ± 0.81 \\
    MiniCheck           & o4-mini      & o4-mini      & $\times$ & 84.09 ± 0.86 & 85.59 ± 0.90 & 82.59 ± 0.81 \\
    MiniCheck  & o4-mini      & -                   & - & 83.95 ± 0.00 & 82.88 ± 0.00 & 85.02 ± 0.00 \\
    
    MiniCheck  & QwQ-32B & QwQ-32B        & $\checkmark$ & 82.25 ± 0.62 & 82.58 ± 1.12 & 81.92 ± 0.83 \\
    MiniCheck           & QwQ-32B & QwQ-32B & $\times$ & 81.72 ± 0.50 & 82.88 ± 1.27 & 80.57 ± 0.33 \\
    MiniCheck  & QwQ-32B & -                   & - & 80.70 ± 0.30 & 82.58 ± 0.42 & 78.81 ± 0.50 \\
    
    \bottomrule
    \end{tabular}
}
\caption{Detailed Results on the WiCE, BioNLI Full and BioNLI-300 dataset.}
\label{tab:our_method_results}
\end{table*}

\begin{figure*}[t]
    \centering
    \begin{subfigure}[t]{0.48\textwidth}
        \centering
        \captionsetup{justification=centering}
        \includegraphics[width=0.9\columnwidth]{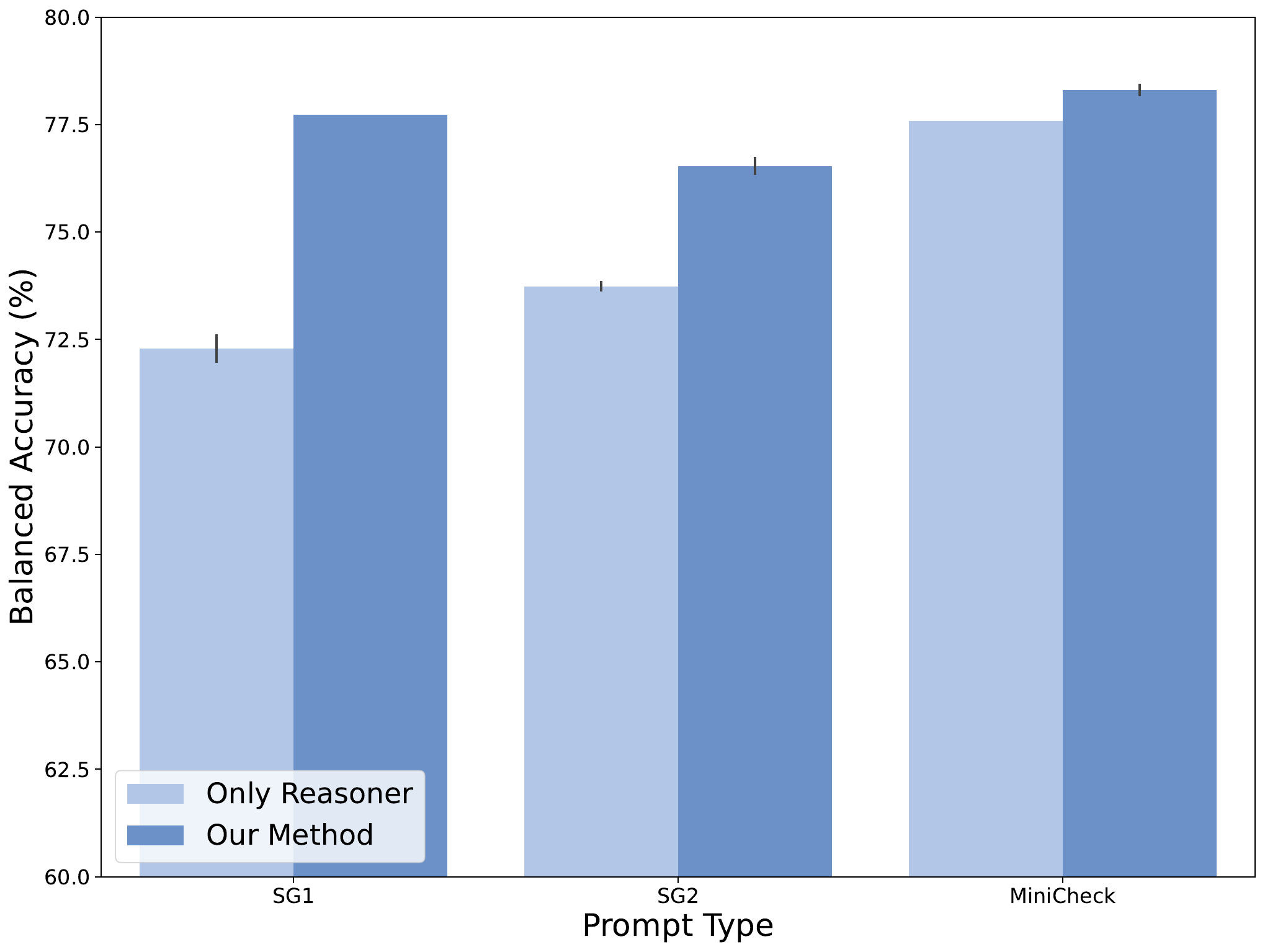}
        \caption{BioNLI dataset with \texttt{Qwen/QwQ-32B} Reasoner and Question-decomposer}
        \label{fig:bionli_a}
    \end{subfigure}%
    ~ 
    \begin{subfigure}[t]{0.48\textwidth}
        \centering
        \captionsetup{justification=centering}
        \includegraphics[width=0.9\columnwidth]{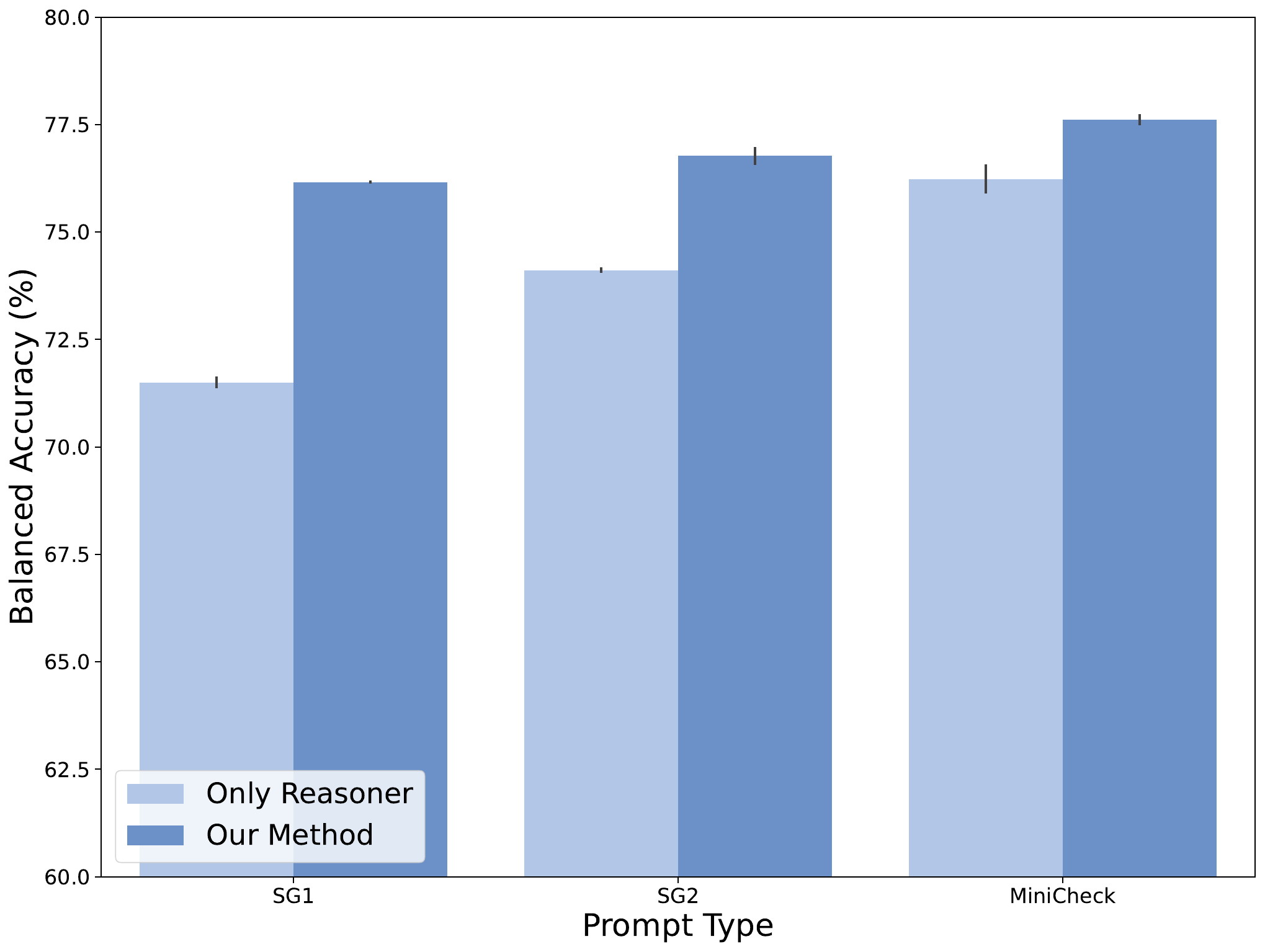}
        \caption{BioNLI dataset with \texttt{Qwen3-32B} Reasoner and Question-decomposer}
        \label{fig:bionli_b}
    \end{subfigure}
    \\ \vspace{1mm}
    \begin{subfigure}[t]{0.48\textwidth}
        \centering
        \captionsetup{justification=centering}
        \includegraphics[width=0.9\columnwidth]{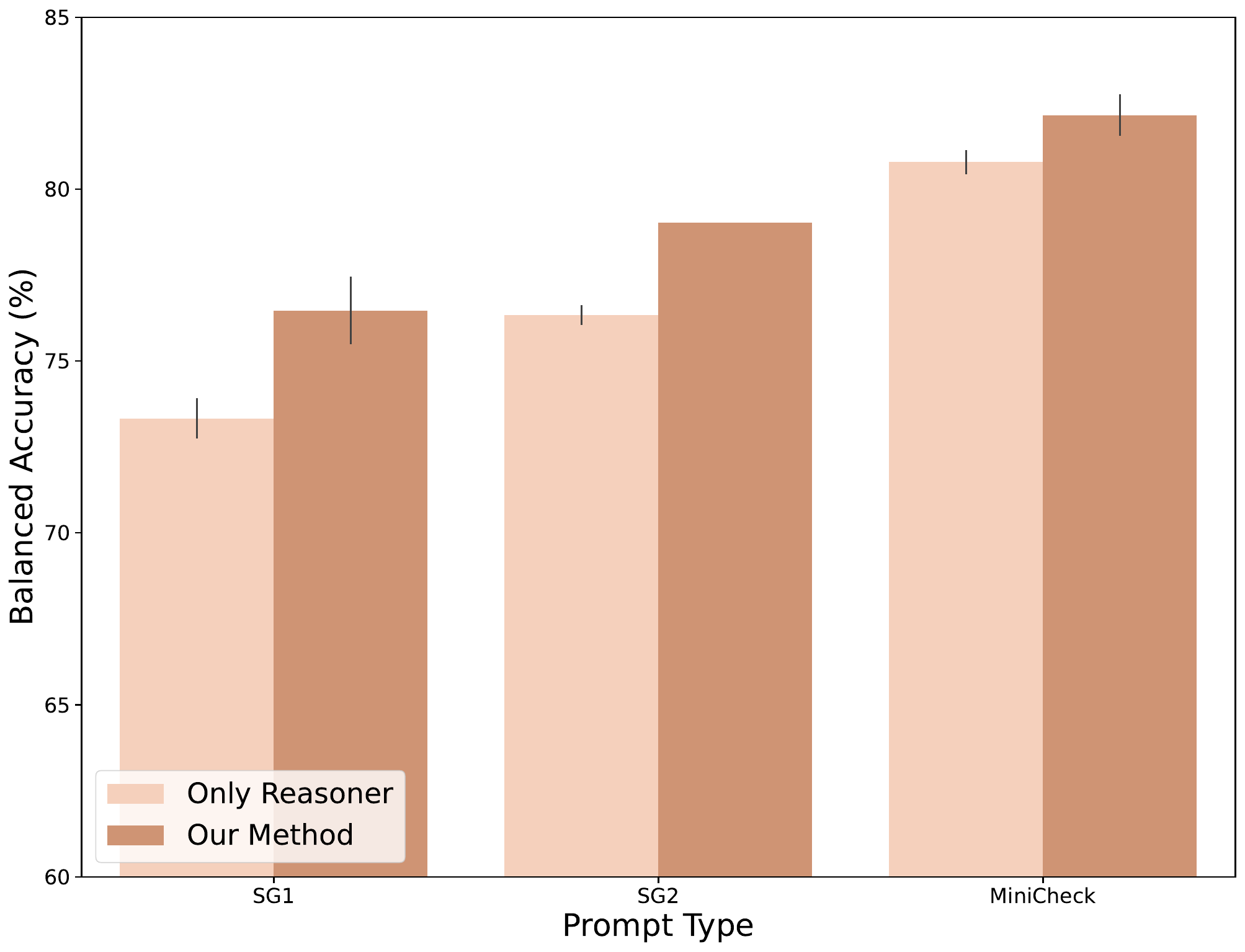}
        \caption{WiCE dataset with \texttt{Qwen/QwQ-32B} Reasoner and Question-decomposer}
        \label{fig:wice_a}
    \end{subfigure}
    ~ 
    \begin{subfigure}[t]{0.48\textwidth}
        \centering
        \captionsetup{justification=centering}
        \includegraphics[width=0.9\columnwidth]{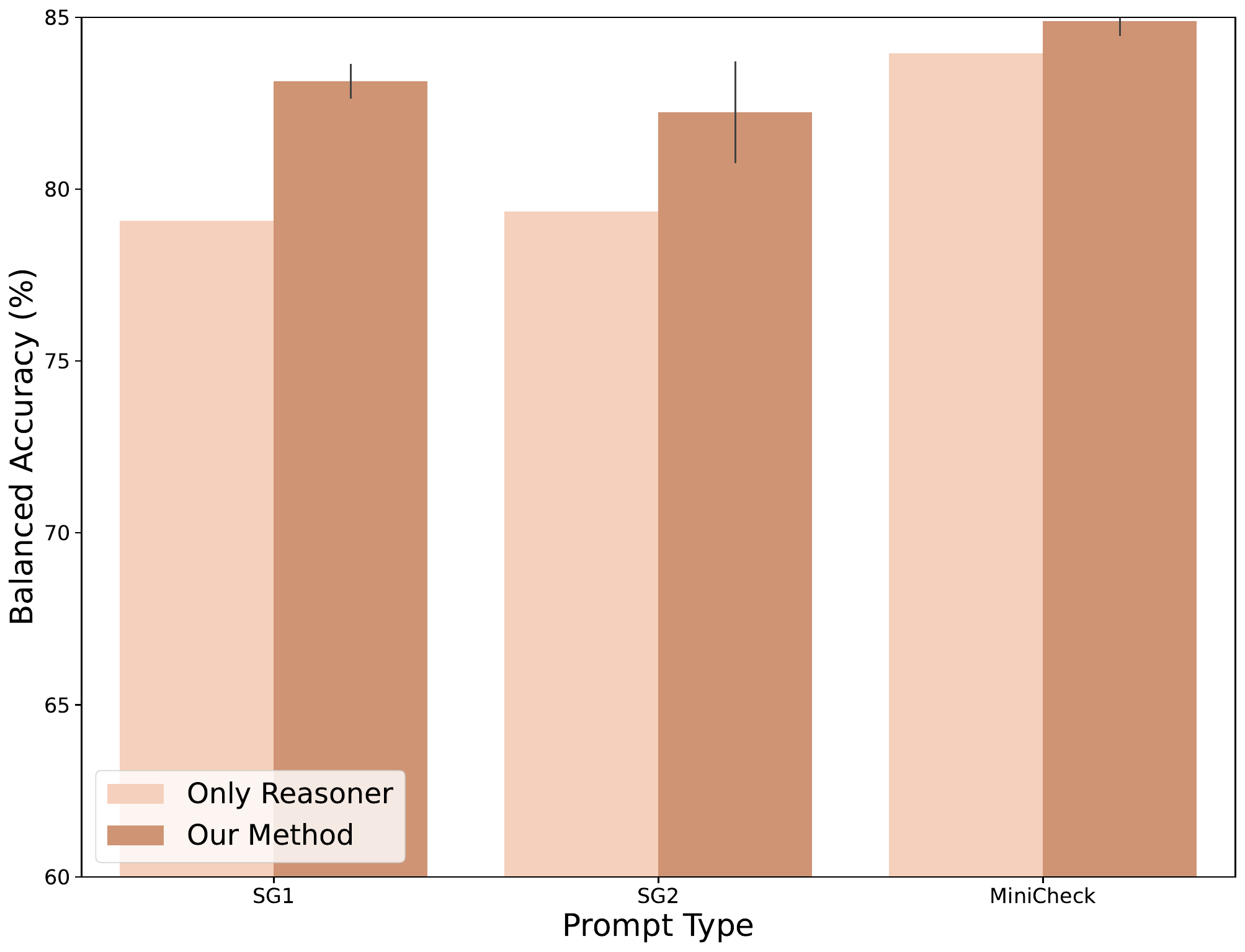}
        \caption{WiCE dataset with \texttt{o4-mini} Reasoner and Question-decomposer}
        \label{fig:wice_b}
    \end{subfigure}
    
    \caption{Balanced accuracy on the BioNLI (\subref{fig:bionli_a}, \subref{fig:bionli_b}) and WiCE (\subref{fig:wice_a}, \subref{fig:wice_b}) benchmarks under three prompt variants. %
    Across each dataset, prompt variant, and backbone LLM, our method yields consistent and significant gains over the single-reasoner baseline. 
    }
    \label{fig:our_method_plot_full}
\end{figure*}

\begin{table}[!t]
    \centering
    \resizebox{\columnwidth}{!}{%
        \begin{tabular}{@{}lccc@{}}
        \toprule
        Prompt    & De-Presupposition & Answer Module & BAcc         \\
        \midrule
        MiniCheck & $\times$          & $\checkmark$  & 71.33 ± 0.00 \\
        MiniCheck & $\times$          & $\times$      & 71.83 ± 0.17 \\
        MiniCheck & $\checkmark$      & $\checkmark$  & 73.50 ± 0.17 \\
        MiniCheck & $\checkmark$      & $\times$      & 74.67 ± 0.33 \\
        \midrule
        SG1       & $\times$          & $\checkmark$  & 69.83 ± 0.83 \\
        SG1       & $\times$          & $\times$      & 71.67 ± 0.00 \\
        SG1       & $\checkmark$      & $\checkmark$  & 73.33 ± 0.67 \\
        SG1       & $\checkmark$      & $\times$      & 73.00 ± 0.33 \\
        \midrule
        SG2       & $\times$          & $\checkmark$  & 69.67 ± 0.67 \\
        SG2       & $\times$          & $\times$      & 71.33 ± 0.33 \\
        SG2       & $\checkmark$      & $\checkmark$  & 71.67 ± 1.00 \\
        SG2       & $\checkmark$      & $\times$      & 74.83 ± 0.50 \\
        \bottomrule
        \end{tabular}%
    }
    \caption{Results (with standard deviation) of using explicit answer module with our method.}
    \label{tab:explicit_answer_module}
\end{table}
\subsection{Qualitative Analysis} \label{app:qual_analysis}
\begin{table*}[]
    \centering
    \small
    \resizebox{0.9\textwidth}{!}{%
    \begin{tabular}{p{0.1\textwidth}|p{0.90\textwidth}}
    \toprule
    \textbf{Label} & \textbf{Content} \\
    \midrule
    
    Claim & We conclude that the reduction in furosemide stimulated renin release by indomethacin is due to renal cyclo-oxygenase inhibition which is reflected in decreased excretion rates of hydrolysis products of renal eicosanoids. \\

    \vspace{0.2mm} \\
    Decomposed Questions & 1: What is the cause of the reduction in furosemide-stimulated renin release?
    
    2: What is the effect of indomethacin on furosemide-stimulated renin release?

    3: What is reflected in decreased excretion rates of hydrolysis products of renal eicosanoids? \\

    \vspace{0.2mm} \\
    Presupposition free Questions & 1: If the reduction in furosemide-stimulated renin release has a cause, what is the cause?
    
2: What is the effect of indomethacin on furosemide-stimulated renin release?

3: If so, has anything been reflected in those decreased excretion rates?

4: What is reflected in decreased excretion rates of hydrolysis products of renal eicosanoids?

5: If indomethacin has an effect on furosemide-stimulated renin release, what is the effect?

6: Does furosemide stimulate renin release?

7: If there is a reduction in furosemide-stimulated renin release, does it have a cause?

8: If yes, what is reflected in them?

9: What is the cause of the reduction in furosemide-stimulated renin release?

10: If furosemide stimulates renin release, does indomethacin have an effect on that release?

11: Do hydrolysis products of renal eicosanoids have excretion rates that can decrease?

12: Is there a reduction in furosemide-stimulated renin release? \\
    
    \vspace{0.2mm} \\
    Reasoning & First, I need to check the document for evidence supporting each part of the claim. 

\textbf{Looking at the document:} The study involved giving indomethacin to normal volunteers and then measuring responses to furosemide. The key points from the document are:

- Indomethacin reduced serum thromboxane B2 (TXB2), which is a platelet product, but that's not directly about the kidney. However, the document mentions that renal prostacyclin and thromboxane A2 synthesis are limited to the cortex. So maybe the eicosanoids in the kidney are affected too.

- The study found that both basal and furosemide-stimulated plasma renin activity were reduced after indomethacin. Also, the excretion rates of 6-keto-PGF1 alpha (a prostacyclin metabolite) and TXB2 (though TXB2 here might be from the kidney, but earlier they said TXB2 is platelet-derived, so maybe the renal TXB2 is different? Wait, the document says "the transient increase in excretion rates of 6-keto-PGF1 alpha and TXB2" were also reduced. So maybe the TXB2 here refers to renal, not platelet? Hmm, that's a bit confusing because earlier they said serum TXB2 was from platelets. Maybe the excretion in urine includes renal sources? The document says "excretion rates" so that's probably urine. So the renal eicosanoids' hydrolysis products would be in urine. 

The claim states that the decrease in renin release is due to cyclo-oxygenase inhibition (which indomethacin does) and this is reflected in decreased excretion of hydrolysis products. The document does mention that after indomethacin, the excretion of 6-keto-PGF1 alpha and TXB2 (the hydrolysis products) decreased, along with the renin. Since prostacyclin and TXA2 are eicosanoids made via cyclo-oxygenase, their hydrolysis products being reduced would indicate that indomethacin (a COX inhibitor) is blocking their synthesis. Thus, the reduced renin release is linked to the inhibition of these eicosanoids, which are part of renal cyclo-oxygenase products. 

\textbf{Looking at the questions:}

1. The cause of the reduction in renin release would be indomethacin's effect via COX inhibition. The document says that after indomethacin, renin is reduced, so that's supported.

2. Indomethacin reduces furosemide-stimulated renin release (document says "reduced, as well as the transient increase..."), so yes.

3. The decreased excretion rates would reflect the inhibition of eicosanoid synthesis. The document mentions decreased excretion of 6-keto-PGF1 and TXB2, so that's the hydrolysis products.

4. The reflection is the renal cyclo-oxygenase inhibition. Since those products are from renal eicosanoids (as per the first sentence about their synthesis being limited to the cortex), their decreased excretion would reflect that inhibition.

5. The effect is reduction of renin release, as in question 2.

6. The document states that furosemide does stimulate renin, as they mention "furosemide stimulated plasma renin activity".

7. The reduction does have a cause (indomethacin's COX inhibition).

8. The decreased excretion rates reflect the inhibition of eicosanoid synthesis.

9. The cause is indomethacin's cyclo-oxygenase inhibition, which the document implies since indomethacin is a COX inhibitor and the eicosanoid products are reduced.

10. Yes, indomethacin reduces the renin release from furosemide.

11. Yes, the document says excretion rates of those products decreased.

12. Yes, the study found a reduction.

\textbf{Putting it all together:} The claim's main points are that the renin reduction is due to COX inhibition (indomethacin) and that this is shown by decreased excretion of hydrolysis products. The document supports both the reduction in renin and the decrease in the excretion of those products (6-keto-PGF1 alpha and TXB2) after indomethacin. Since cyclo-oxygenase inhibition by indomethacin would block the synthesis of prostacyclin and thromboxane, leading to less of their hydrolysis products being excreted, the reasoning in the claim is supported by the document's findings. Therefore, the answer should be \textbf{SUPPORTED}. \\
    \bottomrule
    \end{tabular}
    }
    \caption{One of the examples from the BioNLI dataset where the reasoner correctly verified the claim as \textsc{SUPPORTED}.}
    \label{tab:success}
\end{table*}

\begin{table*}[]
    \centering
    \small
    \resizebox{0.9\textwidth}{!}{%
    \begin{tabular}{p{0.1\textwidth}|p{0.90\textwidth}}
    
    \toprule
    \textbf{Label} & \textbf{Content} \\
    \midrule
    
    Claim & We conclude that ACTH hypersecretion after CRH-mediated stimuli, including naloxone, is an inherent, but variable, feature of DM, caused by expression of the genetic mutation at the anterior pituitary. \\

    \vspace{0.2mm} \\
    Decomposed Questions & 1: Where is the genetic mutation expressed that leads to ACTH hypersecretion in DM?
    
2: What is an example of a CRH-mediated stimulus mentioned in the claim?

3: What causes the ACTH hypersecretion in DM?

4: What type of stimuli trigger ACTH hypersecretion in DM?

5: What is an inherent but variable feature of DM?
\\

    \vspace{0.2mm} \\
    Presupposition-free Questions & 1: If DM exists, does it have any features?
    
2: If there are stimuli that trigger ACTH hypersecretion in DM, what type are they?

3: If DM experiences ACTH hypersecretion, are there any stimuli that trigger it?

4: If ACTH hypersecretion occurs in DM, what causes it?

5: What is an example of a CRH-mediated stimulus mentioned in the claim?

6: If DM has features, does it have any inherent features?

7: What type of stimuli trigger ACTH hypersecretion in DM?

8: If DM exists, does ACTH hypersecretion occur in it?

9: If DM has an inherent but variable feature, which one is it?

10: If DM has inherent features, are any of them variable?

11: What causes the ACTH hypersecretion in DM?

12: Does DM experience ACTH hypersecretion?

13: If such a mutation exists, is it expressed in a specific location?

14: Does DM exist?

15: Where is the genetic mutation expressed that leads to ACTH hypersecretion in DM?

16: If the claim mentions any CRH-mediated stimuli, what is an example?

17: Does a genetic mutation lead to ACTH hypersecretion in DM?

18: If it is expressed in a specific location, where is that?

19: What is an inherent but variable feature of DM?

20: Does the claim mention any CRH-mediated stimuli? \\
    
    \vspace{0.2mm} \\
    Reasoning & First, \textbf{I'll look at the document.} The study shows that in DM patients, their basal ACTH levels are twice as high as controls, and their ACTH responses to naloxone (a CRH stimulator) are four times higher. The cortisol responses were similar, but ACTH was definitely higher. They also mention that 19 out of 36 patients were hyperresponders, and some had reproducible responses. 

\textbf{Now, looking at the questions:}

1. Does DM have features? The document does talk about ACTH hypersecretion as a feature, so yes.
2. The stimuli are CRH-mediated, like naloxone. The document mentions naloxone as a CRH stimulator.
3. Yes, CRH-mediated stimuli trigger ACTH hypersecretion in DM.
4. The cause according to the claim is the genetic mutation in the anterior pituitary. The document says they "previously showed" CRH-mediated stimuli cause this, but does it mention the genetic mutation's role? The document here doesn't explicitly state the genetic mutation's location. Wait, the claim says it's caused by the mutation's expression in the anterior pituitary. The document's current evidence is about the response to naloxone and ACTH levels but doesn't mention the genetic mutation's expression site. Hmm, that's a problem. The study might be part of their previous work, but the current document doesn't specify the cause as the genetic mutation in the anterior pituitary. 

{\color{red} Questions 5-12 are missing}

Looking at the questions 13,14,15,17,18: The claim says the cause is the genetic mutation expressed in the anterior pituitary. The document here doesn't mention the location of the mutation's expression. The study talks about the effect (ACTH levels) but not the exact mechanism's location. So the part about the anterior pituitary might not be supported by the provided document. 

{\color{red} Questions 19,20 are missing}

The claim also says it's an inherent feature. The document does say that DM patients have higher ACTH responses consistently, even in some repeated tests, so that's inherent. The variability is mentioned with hyperresponders (19 out of 36), so variable. 

But the key issue is the cause being the genetic mutation in the anterior pituitary. The document's current text doesn't state that. The first sentence says they "previously showed" CRH-mediated stimuli cause ACTH hypersecretion. Maybe the prior work mentioned the genetic cause, but in the provided document, the cause isn't specified beyond the stimuli leading to the response. The conclusion in the claim adds the genetic mutation's role in the anterior pituitary, which isn't present here. 

Therefore, even though most parts of the claim are supported (CRH-mediated stimuli like naloxone cause ACTH hypersecretion, inherent and variable), the specific part about the genetic mutation's expression in the anterior pituitary isn't addressed in the provided document. So the entire claim \textbf{isn't fully supported} because that part isn't there. \\
    \bottomrule
    \end{tabular}
    }
    \caption{One of the examples from the BioNLI dataset where the reasoner incorrectly verified the claim as \textsc{NOT-SUPPORTED} though the claim is originally \textsc{SUPPORTED}.}
    \label{tab:failure}
\end{table*}

One of the success cases from the BioNLI dataset is shown in \cref{tab:success}, while a failure case is presented in \cref{tab:failure}. The qualitative analysis reveals that, in some instances, the model overlooks portions of the decomposed questions (\cref{tab:failure}), leading to incorrect conclusions despite otherwise relevant evidence.
\subsection{Dataset} \label{app:dataset}
Full dataset statistics of WiCE, BioNLI and BioNLI-300 is provided on \cref{tab:dataset_stat}.

\begin{table*}[!t]
    \centering
        \begin{tabular}{@{}lccccc@{}}
            \multirow{2}{*}{Dataset} & \multirow{2}{*}{Samples} & \multirow{2}{*}{Supported} & \multirow{2}{*}{Refuted} & \multicolumn{2}{c}{\underline{Average \# of Words}} \\
            & & & & Claim & Evidence \\
            \midrule
             WiCE \cite{kwon2023efficient} & 358 & 247 & 111 & 24 & 1316 \\
             BioNLI \cite{bastan_bionli_2022} & 5073 & 3962 & 1111 & 34 & 187 \\
             BioNLI-300 (sampled) & 300 & 150 & 150 & 35 & 185 \\
             FEVER \cite{diggelmann_climate-fever_2021} & 6605 & 3305 & 3300 & 8 & 305 \\
             \bottomrule
        \end{tabular}
    \caption{Statistics of different dataset used in the study. Following previous papers \citep{tang_minicheck_2024, kamoi_wice_2023}, we have converted the WiCE dataset to binary by assuming partially supported as refuted.}
    \label{tab:dataset_stat}
\end{table*}
\subsection{Results on Fever Dataset} \label{app:fever_dataset}
We additionally report results on the FEVER test dataset \citep{Thorne18Fever}. Notably, FEVER was originally developed as a retrieval and coarse verification benchmark rather than a fine-grained verification task. For example, one of the test claims--``Hearts is a musical composition by Minogue''--lacks the complexity necessary for decomposition-based reasoning. Furthermore, as shown in \cref{tab:dataset_stat}, the average claim length is only 8 words, indicating that the claims are already highly atomic and offer limited opportunity for further decomposition.

As shown in \cref{tab:results_fever_dataset}, our decomposition and de-presupposition pipeline does not yield significant improvements on this dataset, highlighting the importance of task complexity needed in benefiting from structured reasoning approaches. Unlike FEVER, most claims in real-world and scientific domains are highly complex, often requiring multi-hop reasoning. This underscores the necessity of decomposition-based approaches for effective claim verification in such settings.

\begin{table*}[!t]
    \resizebox{\textwidth}{!}{%
        \begin{tabular}{@{}lccc>{\columncolor{tablegray}}ccc@{}}
        Prompt              & Reasoner     & Question Decomposer & \thead{De-\\Presupposition} & BAcc         & Supported    & Refuted    \\
        \midrule

        SG2 & Qwen3-32B & Qwen3-32B & $\checkmark$ & 95.27 ± 0.07 & 93.90 ± 0.23 & 96.64 ± 0.09 \\
        SG2 & Qwen3-32B & Qwen3-32B & $\times$     & 95.49 ± 0.11 & 94.55 ± 0.18 & 96.42 ± 0.03 \\
        SG2 & Qwen3-32B &         -             & $\times$     & 95.47 ± 0.04 & 94.49 ± 0.00 & 96.44 ± 0.08 \\
        SG2 & QwQ-32B         & QwQ-32B         & $\checkmark$ & 95.44 ± 0.17 & 94.25 ± 0.24 & 96.62 ± 0.11 \\
        SG2 & QwQ-32B         & QwQ-32B         & $\times$     & 95.42 ± 0.07 & 94.24 ± 0.05 & 96.61 ± 0.09 \\
        SG2 & QwQ-32B         &    -                  & $\times$     & 95.56 ± 0.02 & 94.70 ± 0.09 & 96.42 ± 0.06 \\
        
        SG1 & Qwen3-32B & Qwen3-32B & $\checkmark$ & 95.18 ± 0.17 & 93.92 ± 0.18 & 96.44 ± 0.17 \\
        SG1 & Qwen3-32B & Qwen3-32B & $\times$     & 95.28 ± 0.03 & 94.22 ± 0.06 & 96.33 ± 0.00 \\
        SG1 & Qwen3-32B &          -            & $\times$     & 95.62 ± 0.05 & 95.08 ± 0.08 & 96.17 ± 0.02 \\
        SG1 & QwQ-32B         & QwQ-32B         & $\checkmark$ & 95.31 ± 0.08 & 94.30 ± 0.08 & 96.33 ± 0.09 \\
        SG1 & QwQ-32B         & QwQ-32B         & $\times$     & 95.49 ± 0.06 & 94.58 ± 0.09 & 96.39 ± 0.03 \\
        SG1 & QwQ-32B         &     -                 & $\times$     & 95.64 ± 0.05 & 95.40 ± 0.00 & 95.88 ± 0.09 \\
        
        MiniCheck           & Qwen3-32B & Qwen3-32B & $\checkmark$ & 95.00 ± 0.11 & 93.22 ± 0.21 & 96.77 ± 0.02 \\
        MiniCheck           & Qwen3-32B & Qwen3-32B & $\times$     & 95.36 ± 0.08 & 94.21 ± 0.11 & 96.52 ± 0.06 \\
        MiniCheck           & Qwen3-32B &           -           & $\times$     & 95.41 ± 0.01 & 93.96 ± 0.11 & 96.85 ± 0.09 \\
        MiniCheck           & QwQ-32B         & QwQ-32B         & $\checkmark$ & 95.40 ± 0.02 & 94.13 ± 0.00 & 96.67 ± 0.03 \\
        MiniCheck           & QwQ-32B         & QwQ-32B         & $\times$     & 95.47 ± 0.05 & 94.40 ± 0.03 & 96.53 ± 0.08 \\
        MiniCheck           & QwQ-32B         &     -                 & $\times$     & 95.40 ± 0.03 & 94.16 ± 0.06 & 96.64 ± 0.00 \\
        
        \bottomrule
        \end{tabular}
    }
    \caption{Results (with standard deviation) on the FEVER \citep{Thorne18Fever} dataset. Due to the size of the dataset, we have reported scores over two runs.}
    \label{tab:results_fever_dataset}
\end{table*}

\end{document}